\documentclass[10pt,twocolumn,twoside]{IEEEtran}
\usepackage{cite}

\ifCLASSINFOpdf
\else
\fi
\usepackage{amsmath}
\usepackage{algorithmic}

\usepackage{array}
\usepackage{url}

\usepackage{multirow}
\usepackage{subfigure}
\usepackage{tabularx}
\usepackage{verbatim}
\usepackage{graphicx}
\usepackage{amssymb}
\usepackage{caption2}
\usepackage{balance} 
\usepackage{color}
\usepackage{amsfonts}
\usepackage{algorithm}
\usepackage{algorithmic} 
\begin{document}
%
\title{Fine-grained Visual-textual Representation Learning}
%
%
%

\author{Xiangteng He and Yuxin Peng
\thanks{The authors are with the Institute of Computer Science and Technology,
Peking University, Beijing 100871, China. Corresponding author: Yuxin Peng (e-mail: pengyuxin@pku.edu.cn).}}

\maketitle

\begin{abstract}
Fine-grained visual categorization is to recognize hundreds of subcategories belonging to the same basic-level category, which is a highly challenging task due to the quite subtle and local visual distinctions among similar subcategories. 
Most existing methods generally learn part detectors to discover discriminative regions for better categorization performance. 
However, not all parts are beneficial and indispensable for visual categorization, and the setting of part detector number heavily relies on prior knowledge as well as experimental validation.  
As is known to all, when we describe the object of an image via textual descriptions, we mainly focus on the pivotal characteristics, and rarely pay attention to common characteristics as well as the background areas.
This is an involuntary transfer from human visual attention to textual attention, which leads to the fact that textual attention tells us how many and which parts are discriminative and significant to categorization.
So textual attention could help us to discover visual attention in image.
Inspired by this, we propose a fine-grained visual-textual representation learning (VTRL) approach, and its main contributions are: 
(1) Fine-grained visual-textual pattern mining devotes to discovering discriminative visual-textual pairwise information for boosting categorization performance through jointly modeling vision and text with generative adversarial networks (GANs), which automatically and adaptively discovers discriminative parts. 
(2) Visual-textual representation learning jointly combines visual and textual information, which preserves the intra-modality and inter-modality information to generate complementary fine-grained representation, as well as further improves categorization performance.
Comprehensive experimental results on the widely-used CUB-200-2011 and Oxford Flowers-102 datasets demonstrate the effectiveness of our VTRL approach, which achieves the best categorization accuracy compared with state-of-the-art methods.
\end{abstract}

\begin{IEEEkeywords}
Fine-grained visual categorization, fine-grained visual-textual pattern mining, visual-textual representation learning.
\end{IEEEkeywords}

%
\IEEEpeerreviewmaketitle

\section{Introduction}

\begin{figure}[t]
\begin{center}
\includegraphics[width=1\linewidth]{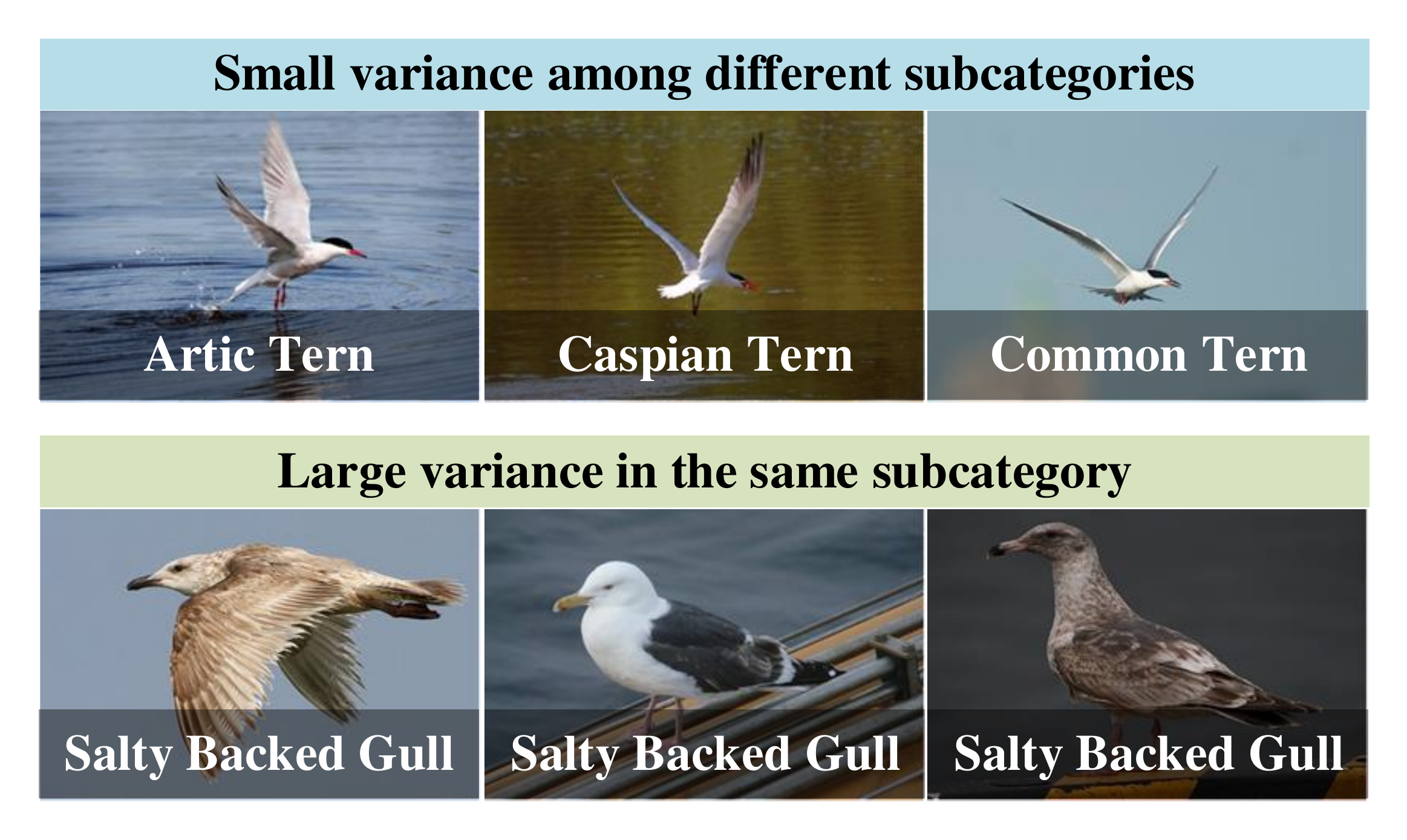}
\end{center}
   \caption{Examples from CUB-200-2011 dataset \cite{cub2011}. Note that fine-grained visual categorization is a technically challenging task even for humans to recognize these subcategories, due to small variances among different subcategories and large variances in the same subcategory.}
\label{example}
\end{figure}

\IEEEPARstart{F}{ine-grained} visual categorization aims to recognize similar subcategories in the same basic-level category. It is one of the most challenging and significant open problems in multimedia and computer vision areas, which has achieved great progress as well as attracted extensive attention of academia and industry in recent years. The progress incarnates in three aspects: (1) More fine-grained domains have been covered, such as animal species \cite{cub2011,stanforddog}, plant breeds \cite{plant,Nilsback08}, car types \cite{krause20133d} and aircraft models \cite{aricraft}. (2) Methodologies of fine-grained visual categorization have achieved promising performance in recent years\cite{huang2016task,zhang2017picking,wang2015deepbag,wang2016branded,He_2017_CVPR}, due to the application of deep neural networks (DNNs). (3) Some information technology companies, such as Microsoft and Baidu, begin to turn fine-grained visual categorization technologies into their applications\footnote{https://www.microsoft.com/en-us/research/project/flowerreco-cn/} \footnote{http://image.baidu.com/?fr=shitu/}.

Fine-grained visual categorization lies in the continuum between basic-level visual categorization (e.g. object recognition) and identification of individuals (e.g. face recognition). 
Its main challenges can be summarized as the following two aspects:
(1) Variances among similar subcategories are subtle and local, because they belong to the same genus. (2) Variances in the same subcategory are large and diverse, due to different poses and views, as well as for animals or plants also because of different living environments and growth periods. 
For example, as shown in Fig. \ref{example}, the images of ``Artic Tern'' and ``Caspian Tern'' look similar in global appearance, but the images of ``Salty Backed Gull'' look different in the pose, view and feather color. So it is hard for a person without professional knowledge to recognize them.

These subcategories can be distinguished by the subtle and local variances of the discriminative parts. It is crucial for fine-grained visual categorization to localize the object and its discriminative parts. 
Researchers generally adopt a two-stage categorization pipeline: the first stage is to localize the object or its discriminative parts, and the second is to extract their features to categorize the subcategory. 
For example, Zhang et al. \cite{zhang2014part} utilize R-CNN \cite{rcnn} with geometric constraints to detect object and its parts first, and then extract the features of the object and its parts, finally train one-versus-all linear SVMs for categorization. 
However, not all the parts are beneficial and indispensable for fine-grained categorization.
The conclusive distinctions among subcategories generally locate at a few specific parts, such as the red beak or the black tail. 
So the categorization performance depends on the number of part detectors and whether the detected parts are discriminative or not. However, mainstream methods generally set the detector number due to their prior knowledge or the experimental validation, which is highly empirical and limited.
For example, when the number of part detectors applied in the experiments increase from eight to fifteen, the performance of fine-grained categorization declines, as reported in \cite{huang2016part}.
Six part detectors are applied by Zhang et al. \cite{picking} to achieve the best categorization accuracy. 
He and Peng \cite{spatialconstraints} applies two discriminative parts for fine-grained categorization.
They are limited in flexibility, and hard to generalize.

Therefore, it is significant to automatically learn how many and which parts really make sense to fine-grained visual categorization. 
When human beings see two images of two different subcategories, human visual attention mechanism plays an important role in focusing on the pivotal distinctions between them. 
Inspired by this, researchers begin to apply human visual attention mechanism in their works, aiming to find the most discriminative characteristics for categorization. Xiao et al. \cite{twoattention} propose a two-level attention model (TL Atten), in which object-level attention selects relevant image proposals to a certain object, and part-level attention selects relevant image proposals to the discriminative parts of the object. Fu et al. \cite{Fu_2017_CVPR} propose a recurrent attention convolutional neural network (RA-CNN) to recursively learn discriminative region attention and region-based feature representation. 
These works simulate human visual attention mechanism to find discriminative parts for categorization from visual information.

\begin{figure}[t]
\begin{center}\includegraphics[width=1\linewidth]{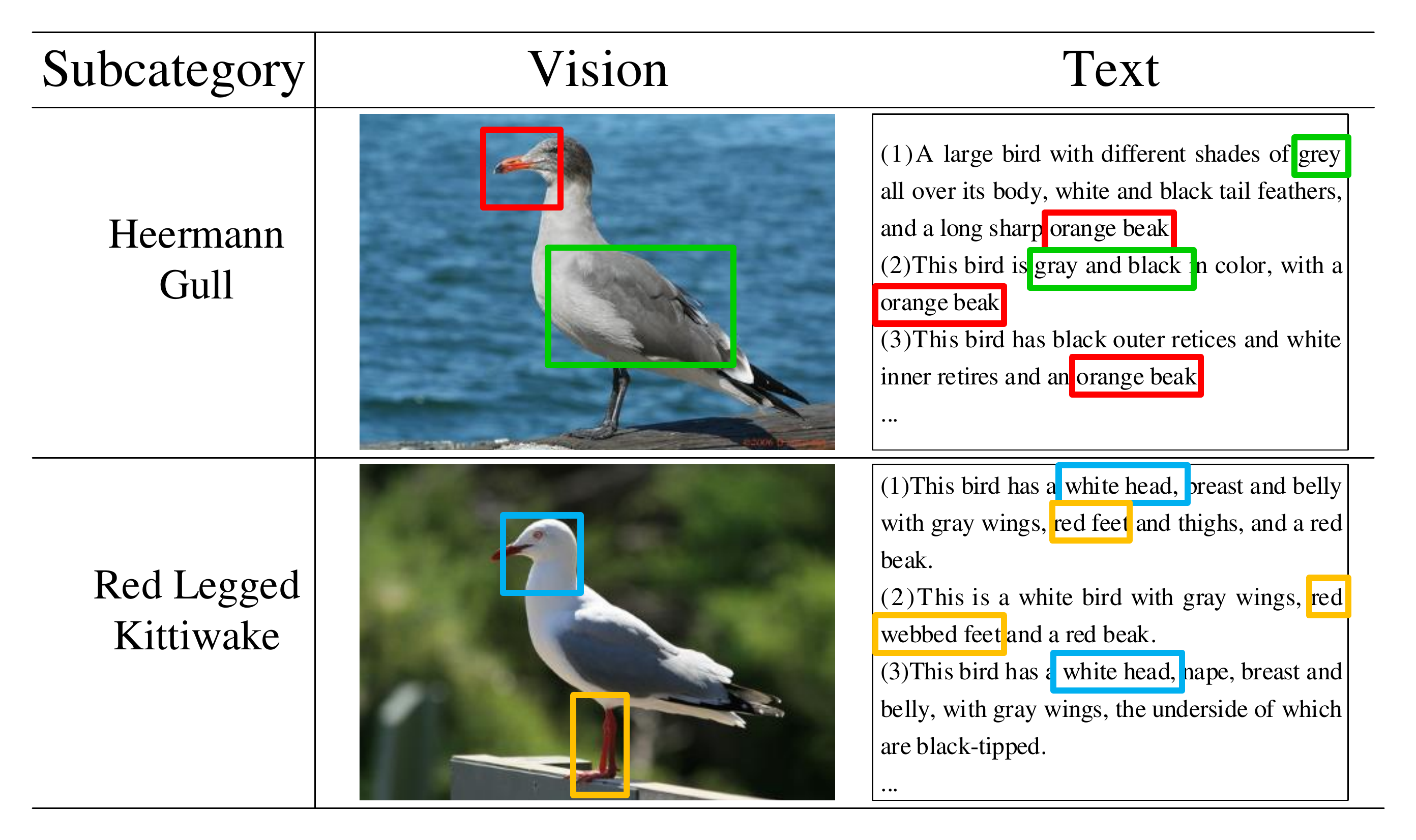}
\caption{Examples of visual and textual attentions. The images come from CUB-200-2011 dataset \cite{cub2011}, and text are collected by Reed et al. \cite{deeprepresentations} through Amazon Mechanical Turk (AMT) platform.}
\label{attentionshow}
\end{center}
\end{figure}

Attention is the behavioral and cognitive process of selectively concentrating on a discrete aspect of information, whether deemed subjective or objective, while ignoring other perceivable information \cite{anderson1985cognitive}. As is known to all, when human beings give the interpretation of the visual data by textual descriptions, they tend to indicate how many and which parts are distinguishing from other subcategories.
These words describing the part attributes are regarded as textual attention, which generally appears frequently in the textual descriptions.
This is an involuntary transfer from human visual attention to textual attention. In this transfer process, common characteristics of object and background areas are ignored naturally.
Textual attention can be obtained by discovering the frequent item sets in the textual descriptions, which point out the discriminative parts of the subcategory. From Fig. \ref{attentionshow}, we can see that the frequent item sets contain ``red break'', which is a discriminative characteristic that distinguishes ``Heermann Gull'' from ``Red Legged Kittiwake''.
\par

Therefore, how to exactly relate textual attention to visual attention and mine the discriminative parts are pivotal to fine-grained visual categorization.
This paper proposes a fine-grained visual-textual representation learning (VTRL) approach, and its main contributions are: 
\begin{itemize}
\item
\textbf{Fine-grained visual-textual pattern mining} devotes to discovering discriminative visual-textual parts for categorization by jointly modeling vision and text with generative adversarial networks (GANs). Different from existing methods, the localized discriminative parts in this paper could not only tell us how many and which parts are significant for categorization, but also which attributes of parts are distinguishing from other subcategories. The part number is determined automatically and adaptively by textual attention. 
\item
\textbf{Visual-textual representation learning} is proposed to combine visual and textual information. Visual stream focuses on the locations of the discriminative parts, while textual stream focuses on the discrimination of the regions. It preserves the intra-modality and inter-modality information to generate complementary fine-grained representation, as well as further improves categorization accuracy.  
\end{itemize}


Our previous conference paper CVL \cite{He_2017_CVPR} proposes a two-stream model combining vision and language for learning the fine-grained representation. Vision stream learns deep representations from visual information and language stream utilizes textual information to encode salient visual aspects for distinguishing subcategories. 
The main differences between the proposed VTRL approach and CVL can be summarized as the following three aspects: 
(1) Our VTRL approach employs textual pattern mining to localize textual attention for exploiting the human visual attention transferred into textual information, which indicates how many and which parts are significant and indispensable for categorization.
While CVL directly utilizes the whole textual information, does not mine fine-grained textual attention information.
(2) Our VTRL approach employs visual pattern mining based on discovered textual patterns to localize discriminative parts, so that discriminative parts and objects are both exploited to learn multi-grained and multi-level representations for boosting fine-grained categorization. 
While CVL only exploits the objects, which ignores the complementary and semantic fine-grained clues provided by the discriminative parts.
(3) Our VTRL approach employs fine-grained visual-textual pattern mining to discover the discriminative and significant visual-textual pairwise information via jointly modeling vision and text with GANs, which  mines the correlation between textual and visual attention.
While CVL only combines vision and text, ignoring to exploit their visual and textual attention, as well as their correlation.
Compared with state-of-the-art methods on two widely-used fine-grained visual categorization datasets, our VTRL approach achieves the best categorization accuracy.

The remainder of this paper is organized as follows: We briefly review the related works in Section \ref{relatedwork}. In Section \ref{approach} our proposed VTRL approach is presented in detail. Then Section \ref{experiments} reports the experimental results and analyses. Finally, Section \ref{conclusion} concludes this paper.

\section{Related Work}
\label{relatedwork}
In this section, we briefly review the related works of fine-grained visual categorization, frequent pattern mining and multi-modal analysis.
\subsection{Fine-grained Visual Categorization}\
Since the discriminative regions of image is crucial for fine-grained visual categorization, most existing methods \cite{zhang2014part,twoattention} first localize the discriminative regions of image, such as the object and its parts, and then extract their discriminative features for fine-grained categorization. 
Some methods directly use the annotations of the object \cite{chai2013symbiotic,yang2012unsupervised} and parts \cite{berg2013poof,xie2013hierarchical} to localize the discriminative regions.
However, it is not available to obtain the annotations in practical applications, some researchers begin to use the annotations of the object and parts only in the training phase. 
Zhang et al. \cite{zhang2013deformable} propose the Deformable Part-based Model (DPM) to localize the discriminative regions with the object and part annotations as the supervised information in the training phase.
Further more, PG Alignment \cite{krause2015fine} is proposed to train part detectors only with object annotation, and localize the discriminative parts in an automatic manner in the testing phase.

Only using object annotation is still not promising in the practical applications.
Recently, some works \cite{twoattention,yao2018autobd,yao2017one} are proposed to localize the discriminative regions in a weakly-supervised manner, which means that neither object nor part annotations are used in both training and testing phases. 
Xiao et al. \cite{twoattention} combine the object and part level attentions to select the discriminative image proposals, which is the first work to localize the discriminative regions without using object and part annotations.
Yao et al. \cite{yao2018autobd} also propose to combine the two complementary object-level and part-level visual descriptions for better performance.
A neural activation constellation (NAC) part model \cite{simon2015neural} is proposed to train part detectors with constellation model.
He and Peng \cite{spatialconstraints} integrate two spatial constraints to select more discriminative proposals and achieve better categorization accuracy.
The aforementioned methods mostly set the detector number due to the prior knowledge or experimental validation, which is highly limited in flexibility and difficult for generalizing to the other domains.
Therefore, we attempt to automatically learn how many and which parts really make sense to categorization via fine-grained visual-textual pattern mining.

\subsection{Frequent Pattern Mining}
Frequent patterns are itemsets, subsequences, or substructures that appear in a data set with frequency no less than a user-specified threshold \cite{han2007frequent}. For example, diaper and beer appear frequently together in sales data of a supermarket, which is a frequent pattern. 
Frequent pattern mining is first proposed by Agrawal et al. \cite{agrawal1993mining} for market basket analysis. Agrawal and Strikant propose Apriori algorithm \cite{agrawal1994fast} to mine frequent patterns in a large transaction database. 
For textual mining, frequent patterns may be sequential patterns, frequent itemsets, or multiple grams. While for visual mining, frequent patterns may be middle-level feature representation or high-level semantic representation. Han et al. \cite{han2000mining} propose to mine visual patterns using low-level features. Li et al.\cite{li2015mid} propose to combine CNN features and association rule mining for discovering visual patterns. Li et al. \cite{li2016event} propose a novel multi-modal pattern mining method, which takes textual pattern and visual pattern into consideration at the same time. 
In this paper, we first utilize Apriori algorithm to discover the textual patterns, and then employ generative adversarial networks (GANs) to mine the relationships between part proposals and textual patterns for better categorization accuracy, which discovers visual and textual patterns at the same time as well as mines the intrinsic correlation between them.

\vspace{-2mm}
\subsection{Multi-modal Analysis}
Nowadays, multi-modal data, e.g. image, text, video and audio, has been widely available on the Internet.
They contains different kinds of information, which are complementary to help achieving comprehensive results in many real-world applications.
So it is significant to learn multi-modal representation for boosting the signal-modal tasks \cite{hong2011beyond,hong2017coherent}.
Canonical correlation analysis (CCA) \cite{hotelling1936relations} is proposed to learn linear projection matrices, which project features of different modalities into the common space and obtain the common representation. It is widely used for modeling multi-modal data \cite{bredin2007audio,hardoon2004canonical,klein2015associating}. 
Zhai et al. propose the joint representation learning method (JRL) to learn projection matrices considering the semantic and correlation information.
Due to the advances of deep learning, deep learning based methods have been proposed to boost the performance of multi-modal representation learning.
Ngiam et al. \cite{ngiam2011multimodal} propose the bimodal autoencoders (Bimodal AE) to model multi-modal data via minimizing the reconstruction error, and learn a shared representation across modalities.

Recently, image and video captioning, which are types of multi-modal analysis, have achieved great progress.
Long Short-Term Memory (LSTM) \cite{hochreiter1997long} and character-based convolutional networks \cite{zhang2015character} are widely used in image and video captioning.
The architecture of Convolutional and Recurrent Networks (CNN-RNN) is widely used in image and video captioning, and achieves great performance. In this paper, we apply the extension of Convolutional and Recurrent Networks (CNN-RNN) to learn a visual semantic embedding. 
In this paper, we bring the multi-modal representation learning into fine-grained visual categorization to jointly modeling vision and text for boosting the performance.

\begin{figure}[!ht]
\begin{center}
\includegraphics[width=1\linewidth]{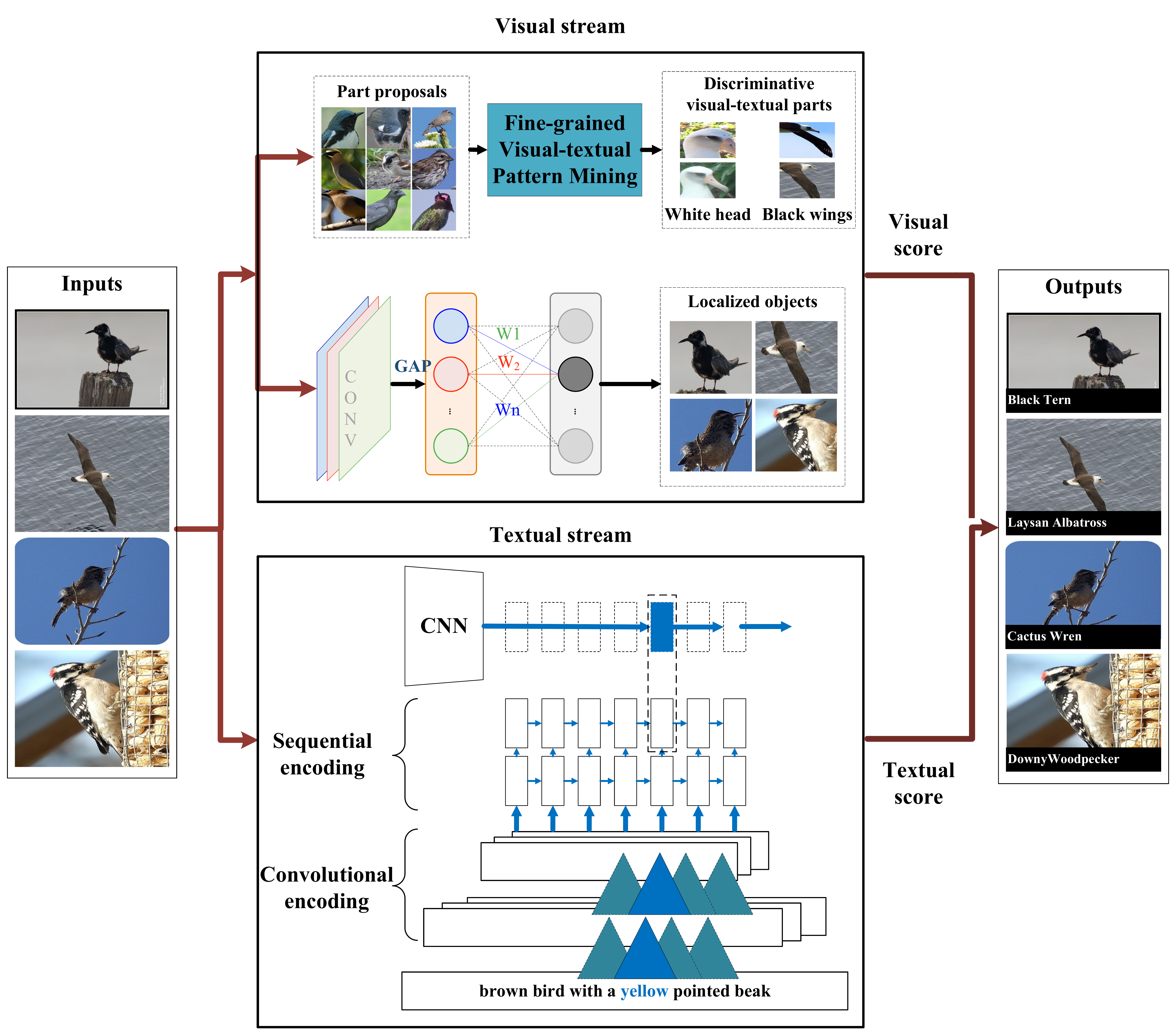}
\end{center}
   \caption{Overview of our VTRL approach.}
\label{framework}
\end{figure}

\section{Our VTRL Approach}
\label{approach}
\subsection{Overview of Our VTRL Approach}
Our approach is based on a very promising and interesting intuition: textual descriptions can point out the discriminative characteristics of images, and provide complementary information with visual information. Therefore, we propose a fine-grained visual-textual representation learning (VTRL) approach, which takes the advantages of visual and textual information jointly as well as exploits the intrinsic correlation between them. Fig. \ref{framework} shows our VTRL approach. First, we conduct fine-grained visual-textual pattern mining to discover the discriminative visual-textual parts as shown in Fig. \ref{textpatternmine}. Then, we localize the object region of image to boost the visual analysis. Finally, we propose a visual-textual representation learning approach to jointly model visual and textual streams for better categorization accuracy.

\subsection{Fine-grained Visual-textual Pattern Mining}
\begin{figure}[t]
\begin{center}
\includegraphics[width=1\linewidth]{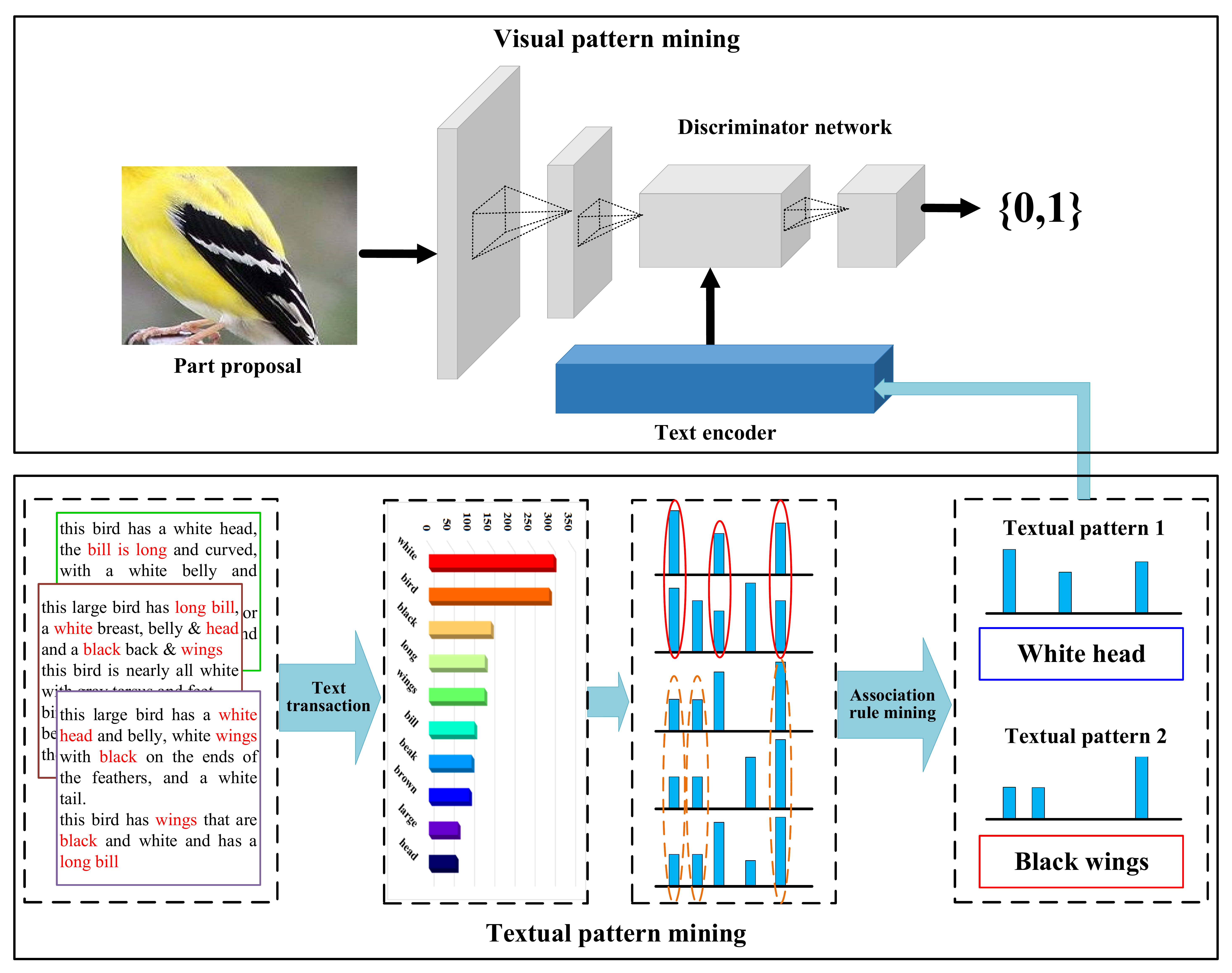}
\end{center}
   \caption{Overview of our fine-grained visual-textual pattern mining approach. $\{0, 1\}$ denotes the output of the discriminator in GANs, which indicates whether the input part proposal meets the input textual pattern.}
\label{textpatternmine}
\end{figure}

Since human visual attention is described into the form of textual descriptions, we first conduct textual pattern mining to discover the textual attention, which indicates the distinguishing part attributes from other subcategories, such as the shape, size and color of the part. Then,
we conduct visual pattern mining to localize the discriminative parts corresponding to the textual patterns discovered by textual pattern mining.  The overview of our fine-grained visual-textual pattern mining approach is shown in Fig. \ref{textpatternmine}.
In the following paragraphs, we describe the fine-grained visual-textual pattern mining approach from three aspects: 1) definition of pattern mining, 2) textual pattern mining and 3) visual pattern mining via GANs.

\subsubsection{Definition of Pattern Mining}
We first introduce the basic definitions for pattern mining. Assume that there is a set of $n$ items, which is denoted as $X=\{x_1, x_2, ..., x_n\}$, and the transaction $T$ is a subset of $X$, i.e. $T \subseteq X$. We also define a transaction database $D=\{T_1, T_2, ..., T_K\}$ that contains $K$ transactions. Our goal is to discover a particular subset $T^*$ of transactions database $X$, which can predict the presence of some target item $y \in T_y$, and $T^* \subset T_y$ as well as $y \cap T^* = \emptyset$. $T^*$ refers to frequent itemset in pattern mining literature. The support of $T^*$ denotes how often $T^*$ appears in $D$ and its definition is as follow:

\begin{gather}
supp(T^*) = \frac{|\{T_y|T^* \subseteq T_y, T_y \in D \}|}{K} 
\end{gather}

An association rule $T^* \to y$ defines a relationship between $T^*$ and a certain item $y$. Therefore, we aim to find patterns that appear in a transaction there is a high likelihood that $y$. We define the confidence as follow: 

\begin{gather}
conf(T^* \to y) = \frac{supp(T^* \cup y)}{supp(T^*)}
\end{gather} 

\subsubsection{Textual Pattern Mining}
In this paper, we devote to discovering textual patterns, which contain the human visual attention information. 
First, we remove stop words and punctuations from each textual description. Then we select the words, which appear in at least 10 textual descriptions in our dataset. Build a vocabulary with these selected words, which is used for generating transactions. It is noted that there are no duplicate words in the vocabulary. In order to generate transaction for each textual description, we map each word back to its corresponding word in the vocabulary, then include that corresponding word index in the transaction. After obtaining the transactions, we perform association rule mining to find the words that frequently appear in textual descriptions, which also means that these words can represent the characteristics of this subcategory. 
Specifically, we utilize the Apriori algorithm \cite{agrawal1994fast} to find a set of patterns $P$ through association rule mining. Each pattern $p \in P$ must satisfy the following criteria:

\begin{gather}
supp(p) > supp_{min} \\
conf(p \to c) > conf_{min}
\end{gather}
where $supp_{min}$ and $conf_{min}$ are thresholds for the support value and confidence value respectively, and $c$ means the image-level subcategory label. After association rule mining, each discovered pattern $p$ contains a set of words. 

We want to find some patterns that point out the discriminative parts of the image, which have the semantic meaning. 
Therefore, we conduct distance constraint on association rule mining as follow:
\begin{gather}
dis(w_i , w_j) < dis_{min}
\end{gather}
where $w_i$ and $w_j$ mean the $i$-th and $j$-th words in the same textual description, and $dis(\cdot)$ means 
the interval between the $i$-th and $j$-th words. The distance function ensures that the discovered patterns have the semantic meaning. The actual threshold in distance function is set to 4 in the experiments, which is set by the cross-validation method following \cite{zhang2014part}.
Finally, we discover a set of patterns $P$, i.e. textual attention in the textual descriptions, which contains the information of human visual attention.

\subsubsection{Visual Pattern Mining via GANs}
After obtaining the textual attention, we devote to mining the relationship between visual and textual attention, i.e. localize the discriminative parts of images via the guidance of textual attention. Due to the great progress made by generative adversarial networks (GANs), which can generate images based on textual information. 
In this paper, we employ GANs to break through the gap between visual and textual information, and localize the discriminative parts corresponding to the discovered textual patterns.
Specifically, the network architecture follows GAN-CLS \cite{reed2016generative}. The original training images and their annotated textual descriptions are used to train the GAN-CLS model. We take the alternating strategy to update the generator and discriminator networks, and use ADAM solver \cite{kingma2014adam} to train the model. The training settings, such as learning rate and momentum, are configured following GAN-CLS \cite{reed2016generative}.

It is noted that part proposals and textual patterns are not used to train the GAN-CLS model, as it is unavailable to obtain their matching labels. Reed et al. \cite{deeprepresentations} point out that the text embedding based on textual descriptions covers the visual attributes, i.e. textual patterns, such as shape, size and color of the part. 
GAN-CLS follows \cite{deeprepresentations} to obtain a visually-discriminative vector representation of text descriptions, by using deep convolutional and recurrent text encoders that learn a correspondence function with images.
Even using images and textual descriptions in the training phase, GAN-CLS can still learn the correlation between the part proposals and textual patterns.
As described in GAN-CLS, the generator has learned to generate plausible images, and also learned to align them with the conditioning information, and likewise the discriminator must learn to evaluate whether samples from generator meet this conditioning constraint. So we first train GAN-CLS on the datasets in our paper, and then apply the discriminator in GAN-CLS to select the corresponding part proposals for the specific textual patterns, where we take one part proposal as the sample input, and one textual pattern as the conditioning constraint input. The selected part proposals contain discriminative information that helps to distinguish similar subcategories.
In the following paragraphs, we introduce the visual pattern mining approach in details.

First, for each image we perform bottom-up process to generate part proposals. In this paper, we utilize selective search method \cite{uijlings2013selective} to generate 1000 part proposals for each image.
Then we take the part proposals and discovered textual patterns as the inputs of discriminator network, to relate the discovered textual patterns with the corresponding part proposals. For each part proposal, discriminator network outputs a score vector that refers to the degree of correlations between part proposal and textual patterns. We select the part proposal with highest score for each textual pattern, which is one of the most discriminative parts for categorization. They will be utilized as the inputs of visual-textual representation learning.  

\subsection{Object Localization}
\label{sec:objectlocalization}
\begin{figure}[t]
\begin{center}
\includegraphics[width=1\linewidth]{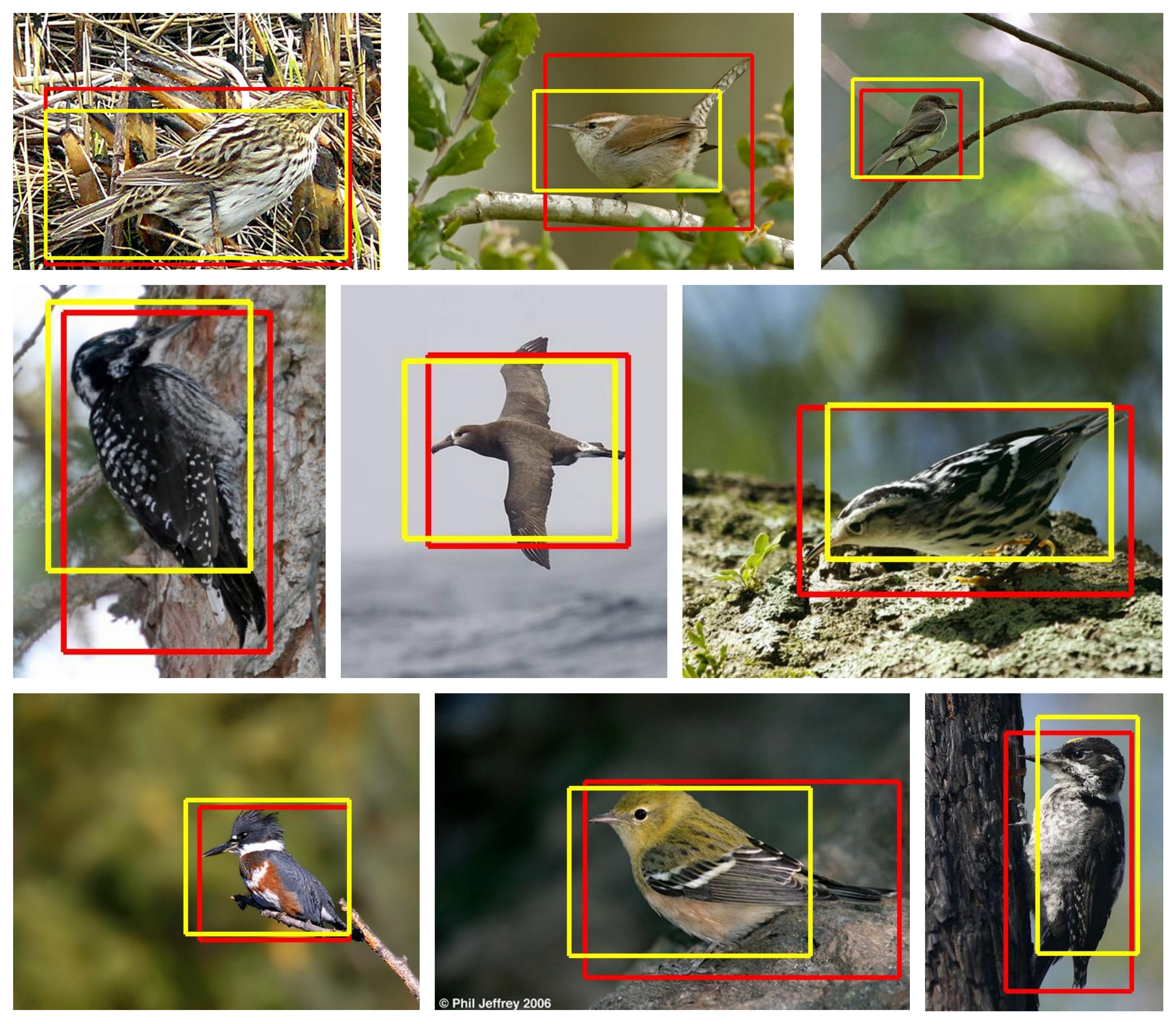}
\end{center}
   \caption{Examples of object localization results in this paper. The red rectangles indicate the ground truth object annotations, i.e. bounding boxes of objects, and the yellow rectangles indicate the object regions localized by our approach.}
\label{cam}
\end{figure}

For better categorization performance, we apply an automatic object localization method based on CAM \cite{zhou2016learning}to localize the object in a weakly-supervised manner, which means that neither object nor part annotations are used in both training and testing phases. Through CAM, we can generate a subcategory activation map $M_c$ for each subcategory $c$, in which the spatial value is calculated as follow:

\begin{gather}
M_c(x,y) = \sum_{k} {w_k^c f_k(x,y)}
\end{gather}
where $f_k(x,y)$ denotes the activation of unit $k$ in the last convolutional layer at spatial location $(x,y)$, and $w_k^c$ is the weight corresponding to subcategory $c$ for unit $k$. The subcategory label information is not available in testing phase, so we set subcategory $c$ by the predicted subcategory. After obtaining the activation map for each image, we conduct OTSU algorithm \cite{otsu1979threshold} to binarize the image and take the bounding box that covers the largest connected area as the localization of object. The localized object is utilized as the inputs of visual-textual representation learning along with the localized discriminative parts via fine-grained visual-textual pattern mining.
Examples of object localization results are shown in Fig. \ref{cam}. It is noted that we use a variant of VGGNet \cite{vgg} as CAM following \cite{zhou2016learning}. In order to get a higher spatial resolution, the layers after conv5\_3 are removed, resulting in a mapping resolution of $14 \times 14$. Besides, a convolutional layer of size $3 \times 3$, stride $1$, pad $1$ with $1024$ neurons is added, followed by a global average pooling layer and a softmax layer.

\subsection{Visual-textual Representation Learning}
Since visual content and textual descriptions provide complementary information, we jointly model them with a two-stream model for learning visual-textual representations to boost the categorization performance. The two-stream model consists of: 1) visual stream and 2) textual stream.

\subsubsection{Visual Stream} 
We apply CNN model, e.g. VGGNet \cite{vgg} in our experiments, as the visual categorization function $f$. The CNN model is pre-trained on the ImageNet 1 K dataset \cite{imagenet}, and then fine-tuned on the fine-grained visual categorization dataset.

For a given image $I$, we first conduct object localization and fine-grained visual-textual pattern mining respectively to obtain the object $b$ and its $n$ discriminative parts $Pa = \{Pa_1, Pa_2, ..., Pa_n\}$.
Then the object and discriminative parts are cropped from the original image, and saved as images $I_b$ and $I_{Pa} = \{I_{Pa_1}, I_{Pa_2}, ..., I_{Pa_n}\}$. We feed the original image $I$ and its object image $I_b$ as well as its part images $I_{Pa} = \{I_{Pa_1}, I_{Pa_2}, ..., I_{Pa_n}\}$ to the CNN model to obtain the predicted visual scores. For the part images, we calculate their mean value as the final part prediction. Finally, we calculate the weighted mean of original prediction, object prediction and part prediction as the final visual prediction.

\subsubsection{Textual Stream} 
In textual stream, we aim to measure the similarity between visual and textual information.
We first apply the deep structured joint embedding method \cite{deeprepresentations} to jointly embed vision (i.e. images) and text (i.e. natural language descriptions for images), which learns a compatibility function of vision and text. 
\par
We define the training data as $D={(v_n,t_n,y_n), n=1, ..., N}$, where $v \in V$ and $t \in T$ denote the vision and text, and $y \in Y$ denotes their subcategory labels.
Then we apply the empirical risk to learn the visual and textual classifier functions $f_v : V \to Y$ and $f_t : T \to Y$ as follows:

\begin{gather}
\frac{1}{N} \sum_{n=1}^N \Delta (y_n,f_v(v_n))+ \Delta (y_n,f_t(t_n))
\end{gather}
where $\Delta : y \times y \to \mathbb{R}$ is the 0-1 loss and
\begin{gather}
f_v(v)=arg \max_{y \in Y} \mathbb{E}_{t \thicksim T(y)}[F(v,t)] \\
f_t(t)=arg \max_{y \in Y} \mathbb{E}_{v \thicksim V(y)}[F(v,t)]
\end{gather}

The compatibility function $F : V \times Y \to \mathbb{R}$ is defined as the inner product of features from the learnable encoder functions as follows:
\begin{gather}
F(v,t)= \theta (v)^T \phi(t)
\end{gather}
where $\theta (v)$ is the visual encoder, and $\phi (t)$ is the textual encoder.
The visual and textual encoders are implemented by GoogleNet \cite{googlenet} and Convolutional Recurrent Net (CNN-RNN) \cite{deeprepresentations} respectively in our approach.
The CNN-RNN model consist of a mid-level temporal CNN hidden layer and a recurrent network.
The outputs of the hidden unit over the textual sequence is averaged as the textural features. Then the textual predicted score is defined as a linear accumulation of evidence for compatibility with the image which needs to be recognized.

\subsection{Training Process}
In this subsection, we summarize our training process.
We train three models for original images, objects and parts respectively. Their detailed training processes are shown in Algorithm \ref{trainingprocess}. 

\begin{algorithm}
  \caption{Training Process}
  \label{trainingprocess}
  \begin{algorithmic}[1]
    \REQUIRE  
    The training images $I$ and their corresponding textual descriptions $T$.
    \ENSURE The model $M$.
    \STATE Set $M=\{M_{ori}, M_{object}, M_{part}\}$
    \STATE Use $I$ to fine-tune the CNN model, which is pre-trained on ImageNet, obtaining the model $M_{ori}$
    \STATE Conduct object localization as described in Section \ref{sec:objectlocalization}, to get the object regions $b$ of $I$
    \STATE Crop $b$ from $I$ and save as images $I_b$
    \STATE Use $I_b$ to fine-tune $M_{ori}$, obtaining the model $M_{object}$
    \STATE Follow \cite{reed2016generative} to train GAN-CLS using minibatch SGD with $I$ and $T$ as pairwise constraints
    \STATE Conduct selective search \cite{uijlings2013selective} on each image to get part proposals $S$
    \STATE Conduct textual pattern mining to obtain the discriminative textual patterns $P$ for each subcategory
    \FOR{$k = 1,...,n; j=1,...,d$}
    \STATE Take $k$-th part proposal $S_k$ and $j$-th textual pattern $P_j$ as the input of the generator $\mathcal{G}$ of GAN-CLS
    \STATE Perform a feed-forward pass, and output the correlation score of $S_k$ and $P_j$
    \STATE For $P_j$ we select one part proposal with the highest correlation score
    \ENDFOR
    \STATE Use the selected part proposals to  fine-tune $M_{object}$, obtaining the model $M_{part}$
    \RETURN $M$.
  \end{algorithmic}
\end{algorithm}

\subsection{Final Prediction}
For a given image $I$, we obtain the visual predicted score from the view of the visual information, and obtain the textual predicted score via measuring the visual and textual information with the shared compatibility function. Due to the fact that joint learning of visual and textual information preserves the intra-modality and inter-modality information to generate complementary information, we fuse the visual and textual predicted results as the final prediction via the follow equation:
\begin{gather}
f(I)= f_v(v) + \beta * f_t(t)
\end{gather}
where $f_v(v)$ and $f_t(t)$ are the visual and textual predicted scores as mentioned above. $\beta$ is selected by the cross-validation method following \cite{zhang2014part}, and its value is $2$ in our experiments on the two fine-grained datasets.  

\begin{figure*}[t]
\begin{center}
\includegraphics[width=1\linewidth]{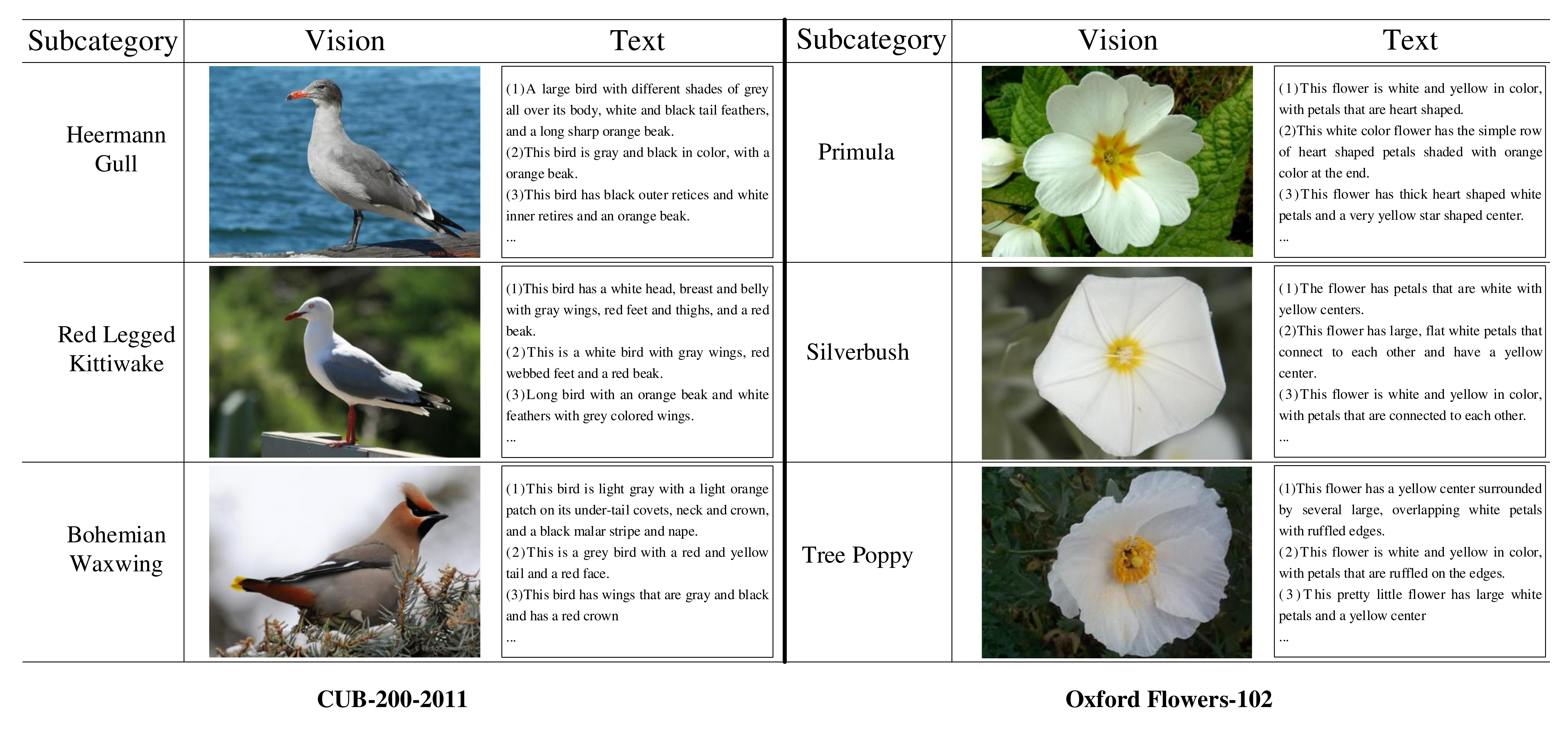}
\end{center}
   \caption{Some examples of vision and text in CUB-200-2011 dataset and Oxford Flowers-102 dataset.}
\label{language example}
\end{figure*}

\section{Experiments}
\label{experiments}
\subsection{Datasets}
This subsection presents two fine-grained visual categorization datasets adopted in the experiments, including CUB-200-2011 and Oxford Flowers-102 datasets, and their detailed information is described as follows:
\begin{itemize} 
\item
\textbf{CUB-200-2011}. It is the most widely-used dataset for fine-grained visual categorization task. The visual information comes from the original dataset of CUB-200-2011 \cite{cub2011}. It contains 11,788 images of 200 subcategories belonging to birds, 5,994 for training and 5,794 for testing. Each image has detailed annotations: 1 subcategory label, 15 part locations, 312 binary attributes and 1 bounding box. The textual information comes from Reed et al. \cite{deeprepresentations}. They expand the CUB-200-2011 dataset by collecting fine-grained natural language descriptions. Ten single-sentence descriptions are collected for each image, as shown in Fig. \ref{language example}. The natural language descriptions are collected through the Amazon Mechanical Turk (AMT) platform, and are required at least 10 words, without any information of subcategories and actions. 

\item
\textbf{Oxford Flowers-102}. Same with CUB-200-2011 dataset, textual information comes from Reed et al. \cite{deeprepresentations}, and visual information comes from the original dataset of Oxford Flowers-102 \cite{Nilsback08}, as shown in Fig. \ref{language example}. It has 8,189 images of 102 subcategories belonging to flowers, 1,020 for training, 1,020 for validation and 6,149 for testing. Each subcategory consists of between 40 and 258 images.
\end{itemize}

\begin{figure*}[t]
\begin{center}
\includegraphics[width=1\linewidth]{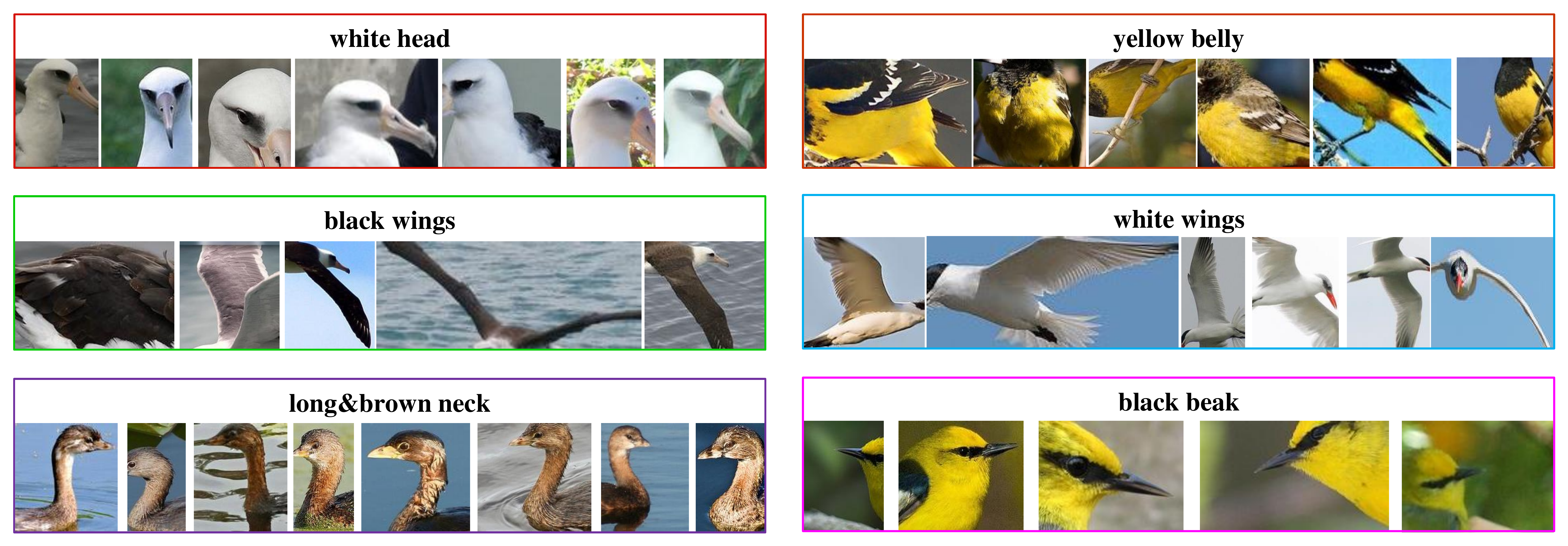}
\end{center}
   \caption{Examples of the matching between textual patterns and visual patterns in our fine-grained visual-textual pattern mining approach.}
\label{patternexample}
\end{figure*}
\subsection{Evaluation Metric}
\textbf{Accuracy} is adopted to comprehensively evaluate the categorization performances of our VTRL approach as well as compared state-of-the-art methods, which is widely used in fine-grained visual categorization \cite{zhang2014part,zhang2017picking}, and its definition is as follow:
\begin{gather}
Accuracy = \frac{R_a}{R}
\end{gather} 
where $R$ denotes the number of images in testing set, and $R_a$ denotes the number of images that are correctly classified.

\begin{table*}[!ht]
   \centering
    \caption{Comparisons with state-of-the-art methods on CUB-200-2011, sorted by amount of annotation used. ``Bbox'' and ``Parts''indicate the object and part annotations (i.e. bounding box and parts locations) provided by the dataset.}
   \begin{tabular} {|p{4cm}<{\centering}|p{1cm}<{\centering}|p{1cm}<{\centering}|p{1cm}<{\centering}|p{1cm}<{\centering}|p{2cm}<{\centering}|}
      \hline
      \multirow {2}{*}{Method} & \multicolumn{2}{|c|}{Train Annotation} & \multicolumn{2}{|c|}{Test Annotation} & \multirow {2}{*}{Accuracy (\%)} \\ 
      \cline{2-5}
      &Bbox & Parts & Bbox & Parts &\\
      \hline
      \hline
      \textbf{Our VTRL Approach} & & & & & {\textbf{86.31}} \\
      \hline
      Fused CN-Nets \cite{yao2017one} & & & & & 85.65 \\
      CVL \cite{He_2017_CVPR} & & & & & 85.55 \\
      RA-CNN \cite{Fu_2017_CVPR} & & & & & 85.30 \\
      PNA \cite{zhang2017picking} & & & & & 84.70 \\
      TSC \cite{spatialconstraints} & & & & & 84.69 \\
      FOAF \cite{zhang2016fused} & & & & & 84.63 \\
      Low-rank Bilinear \cite{Kong_2017_CVPR} & & & & & 84.21 \\
      Spatial Transformer \cite{jaderberg2015spatial} & & & & & 84.10 \\
      Bilinear-CNN \cite{lin2015bilinear} & & & & & 84.10 \\
      Multi-grained \cite{wang2015multiple} & & & & & 81.70 \\
      AutoBD \cite{yao2018autobd} & & & & & 81.60 \\
      NAC \cite{simon2015neural} & & & & & 81.01  \\
      PIR \cite{zhang2016weakly}& & & & & 79.34 \\
      TL Atten \cite{twoattention} & & & & & 77.90  \\
      MIL \cite{xu2017friend} & & & & & 77.40 \\
      VGG-BGLm \cite{zhou2015fine} & & & & & 75.90  \\
      Dense Graph Mining \cite{zhang2016detecting} & & & & & 60.19 \\
      \hline
      Coarse-to-Fine \cite{yao2016coarse}& $\surd$ & & & & 82.50 \\
      PG Alignment \cite{krause2015fine} & $\surd$ & & $\surd$ & & 82.80 \\
      Triplet-A (64) \cite{cui2015fine} & $\surd$ & & $\surd$ & & 80.70\\
      \hline
      Webly-supervised \cite{xu2016webly} & $\surd$ & $\surd$ &  &  & 78.60 \\
      PN-CNN \cite{branson2014bird} & $\surd$ & $\surd$ &  &  & 75.70 \\
      Part-based R-CNN \cite{zhang2014part} & $\surd$ & $\surd$ &  & & 73.50 \\
      SPDA-CNN \cite{zhangspda} & $\surd$ & $\surd$ &$\surd$ &  & 85.14\\
      Deep LAC \cite{lin2015deep} & $\surd$ & $\surd$ & $\surd$ &  & 84.10 \\
      PBC \cite{huang2017pbc} & $\surd$ & $\surd$ & $\surd$ &  & 83.70 \\
      SPDA-CNN \cite{zhang2016spda} & $\surd$ & $\surd$ & $\surd$ &  & 81.01 \\
      PS-CNN \cite{huang2016part}& $\surd$ & $\surd$ & $\surd$ &  & 76.20 \\
      PN-CNN \cite{branson2014bird} & $\surd$ & $\surd$ & $\surd$ & $\surd$ & 85.40  \\
      \hline
   \end{tabular}
   \label{cubresult}
\end{table*}

\begin{table}
   \caption{Comparisons with state-of-the-art methods on Oxford Flowers-102.}
   \begin{center}
   \begin{tabular} {|p{5cm}<{\centering}|p{2cm}<{\centering}|}
      \hline
      Method & Accuracy (\%)\\ 
      \hline
      \hline
      \textbf{Our VTRL Approach} & \textbf{96.89}\\ 
      \hline
      CVL \cite{He_2017_CVPR} & 96.21 \\
      PBC \cite{huang2017pbc} & 96.10 \\
      NAC \cite{simon2015neural}& 95.34 \\
      RIIR \cite{xie2017towards} & 94.01 \\
      Deep Optimized \cite{azizpour2015generic} & 91.30 \\
      SDR \cite{azizpour2015generic}& 90.50 \\
      MML \cite{qian2015fine} & 89.45 \\
      CNN Feature \cite{sharif2014cnn} & 86.80 \\
      Generalized Max Pooling \cite{murray2014generalized} & 84.60 \\  
      Efficient Object Detection \cite{plant} & 80.66 \\
      \hline
   \end{tabular}
   \end{center}
   \label{flowerexperiment}
\end{table}

\subsection{Implementation Details}
\noindent{\textbf{Fine-grained Visual-textual Pattern Mining.}} First, we calculate the frequency of each word in the textual descriptions for each subcategory, and select the top-10 words as keywords, and then discover textual frequent patterns via Apriori algorithm \cite{agrawal1994fast}.
It is noted that we conduct textual pattern mining for each subcategory respectively rather than all subcategories together, which guarantees that the frequent textual patterns tend to be the descriptions of the discriminative parts, such as ``white head'', ``black wings'' and ``long bill''. Second, we conduct selective search \cite{uijlings2013selective} on each image to generate part proposals. Finally, we employ discriminator network to relate textual patterns to part proposals, then select the proposal with highest score as the discriminative part for each textual pattern. For each subcategory, the number of parts is set automatically and adaptively based on the discovered textual patterns.
Fig. \ref{patternexample} shows some matching examples between textual pattern and visual pattern, which are the discriminative characteristics of the subcategory, such as ``long\&brown neck'', ``yellow belly'' and ``black beak''.

\noindent{\textbf{Visual-textual Representation learning.}} For textual stream, we apply CNN-RNN \cite{deeprepresentations} as the text encoder to learn a correspondence function with images. In the training phase, we follow Reed et al. \cite{deeprepresentations}. 
For visual stream, we apply the widely-used model 19-layer VGGNet \cite{vgg} with batch normalization. The model is first pre-trained on ImageNet 1K dataset, and then fine-tuned on the fine-grained visual categorization dataset. Inspired by the strategy adopted by Xiao et al \cite{twoattention}, we utilize the pre-trained CNN model as a filter net to select proposals relevant to the object from the generated image proposals by selective search method. We further fine-tune the pre-trained model with the selected image proposals.

\subsection{Comparisons with state-of-the-art methods}
In this subsection, we present the experimental results of our proposed approach as well as all the compared state-of-the-art methods, as shown in Tables \ref{cubresult} and \ref{flowerexperiment}, which demonstrate the effectiveness of our proposed VTRL approach. As shown in Table \ref{cubresult}, our proposed VTRL approach improves the categorization accuracy from 85.65\% to 86.31\% on CUB-200-2011 dataset. We divide the compared methods into three groups due to the usage of object and part annotations in these methods.
\begin{itemize} 
\item
\emph{Neither object nor part annotations are used.} Nowadays, researchers focus on how to get better categorization accuracy under the weakly-supervised setting, which means neither object nor part annotations are used. Most of these methods utilize the attention property of convolutional neural layers to localize the discriminative parts of object for better accuracy, such as Fused CN-Nets \cite{yao2017one}, RA-CNN \cite{Fu_2017_CVPR}, PNA \cite{zhang2017picking}, TSC \cite{spatialconstraints} and TL Atten \cite{twoattention}. They simulate human visual attention mechanism only from visual information. In our approach, we exploit visual and textual attention simultaneously as well as mine the complementary information between them, which make our proposed approach more effective and obtain a 0.66\% higher accuracy than the best performing result of Fused CN-Nets \cite{yao2017one}.
We also compare with our previous conference work, i.e. CVL \cite{He_2017_CVPR}. We can see that our VTRL approach brings improvements than CVL by 0.76\% and 0.67\% respectively on CUB-200-2011 and Oxford Flowers-102 datasets. It is mainly because that the VTRL approach exploits the textual attention to localize discriminative regions, while CVL directly uses the whole textual descriptions and does not consider the discriminative regions in the images.
\item
\emph{Only one of object and part annotations is used.} These methods utilize object annotation (i.e. bounding box) to train an object detector or learn part detectors, which are to learn more representative features for categorization. In our approach, we utilize CAM \cite{zhou2016learning} to automatically localize the object region of image, which avoids using object annotation.
The result of object localization can be seen in Fig. \ref{attentionshow}. Even using object annotation, these methods achieve lower accuracies than our proposed VTRL approach.
\item
\emph{Both object and part annotations are used.} In order to obtain better categorization accuracy, some methods utilize both object and part annotations at training phase as well as testing phase. However, these annotations are heavy labor-consuming. In our approach, we get object region and discriminative parts automatically via object localization and fine-grained visual-textual pattern mining respectively without using any annotations. We promote the categorization performance through discovering the discriminative and representative object and its parts.   
\end{itemize}

Besides, categorization results on Oxford Flowers-102 dataset are shown in Table \ref{flowerexperiment}, and also have the similar trend as CUB-200-2011 dataset, while our proposed VTRL approach still keeps the best.

\subsection{Effects of Components in Our VTRL Approach}

In this subsection, we conduct two baseline experiments to verify the separate contribution of each component in our proposed VTRL approach. Tables \ref{vision} to \ref{cvl} show the accuracies of our proposed VTRL approach as well as the baseline approaches on CUB-200-2011 dataset at the following two aspects.

\begin{table}
\caption{Effects of fine-grained pattern mining and object localization for visual stream. }
   \begin{center}
   \begin{tabular} {|p{5cm}<{\centering}|p{2cm}<{\centering}|}
      \hline
      Method & Accuracy (\%)\\ 
      \hline
      \hline
      \textbf{VTRL-visual} & \textbf{85.54}\\
      \hline
      VTRL-visual(w/o object) & 83.21 \\
      \hline
      VTRL-visual(w/o parts) & 84.79 \\
      \hline
      VTRL-visual(w/o object\&parts) & 80.82 \\
      \hline
   \end{tabular}
   \end{center}
   \label{vision}
\end{table}

\begin{table}
\caption{Effects of different components of our proposed approach on CUB-200-2011. }
   \begin{center}
   \begin{tabular} {|p{5cm}<{\centering}|p{2cm}<{\centering}|}
      \hline
      Method & Accuracy (\%) \\ 
      \hline
      \hline
      \textbf{Our VTRL Approach} & \textbf{86.31} \\
      \hline
      VTRL(w/o object) & 85.17 \\
      \hline
      VTRL(w/o parts) & 85.83 \\
      \hline
      VTRL(w/o object\&parts) & 84.05 \\
      \hline
   \end{tabular}
   \end{center}
   \label{vta}
\end{table}

\subsubsection{Effects of Fine-grained Visual-textual Pattern Mining and Object Localization}
In our VTRL approach, fine-grained visual-textual pattern mining and object localization generate discriminative parts and object for promoting the categorization accuracy. They make sense to the visual stream and then further impact whole approach. Tables \ref{vision} and \ref{vta} show the effects of fine-grained visual-textual pattern mining and object localization to visual stream and our proposed VTRL approach respectively. In the tables, ``object'' means that object localization is conducted, and ``parts'' means that fine-grained visual-textual pattern mining is employed. We can observe that considering object localization can achieve better categorization accuracy than considering fine-grained visual-textual pattern mining. This is because that objects contain the global and local features simultaneously, while discriminative parts focus subtle and local characteristics. However, jointly considering object localization and fine-grained visual-textual pattern mining can further improve the categorization accuracy. 

Fine-grained visual-textual pattern mining aims to select the part proposals that corresponding to the discovered textual patterns. The relations between part proposals and textual patterns ensure the discrimination and representativeness of selected parts. Some examples of discovered visual-textual patterns are shown in Fig. \ref{patternexample}.

\begin{table}
\caption{Effects of different components of our proposed approach on CUB-200-2011.}
   \begin{center}
   \begin{tabular} {|p{5cm}<{\centering}|p{2cm}<{\centering}|}
      \hline
      Method & Accuracy (\%)\\ 
      \hline
      \hline 
      \textbf{Our VTRL Approach} & \textbf{86.31} \\
      \hline
      VTRL-textual & 81.81 \\
      \hline
      VTRL-visual & 85.54 \\
      \hline
      VTRL(only original image) & 80.82 \\
      Co-attention \cite{lu2016hierarchical} & 73.90 \\
      \hline
   \end{tabular}
   \end{center}
   \label{cvl}
\end{table}
\begin{figure}[t]
\begin{center}
\includegraphics[width=1\linewidth]{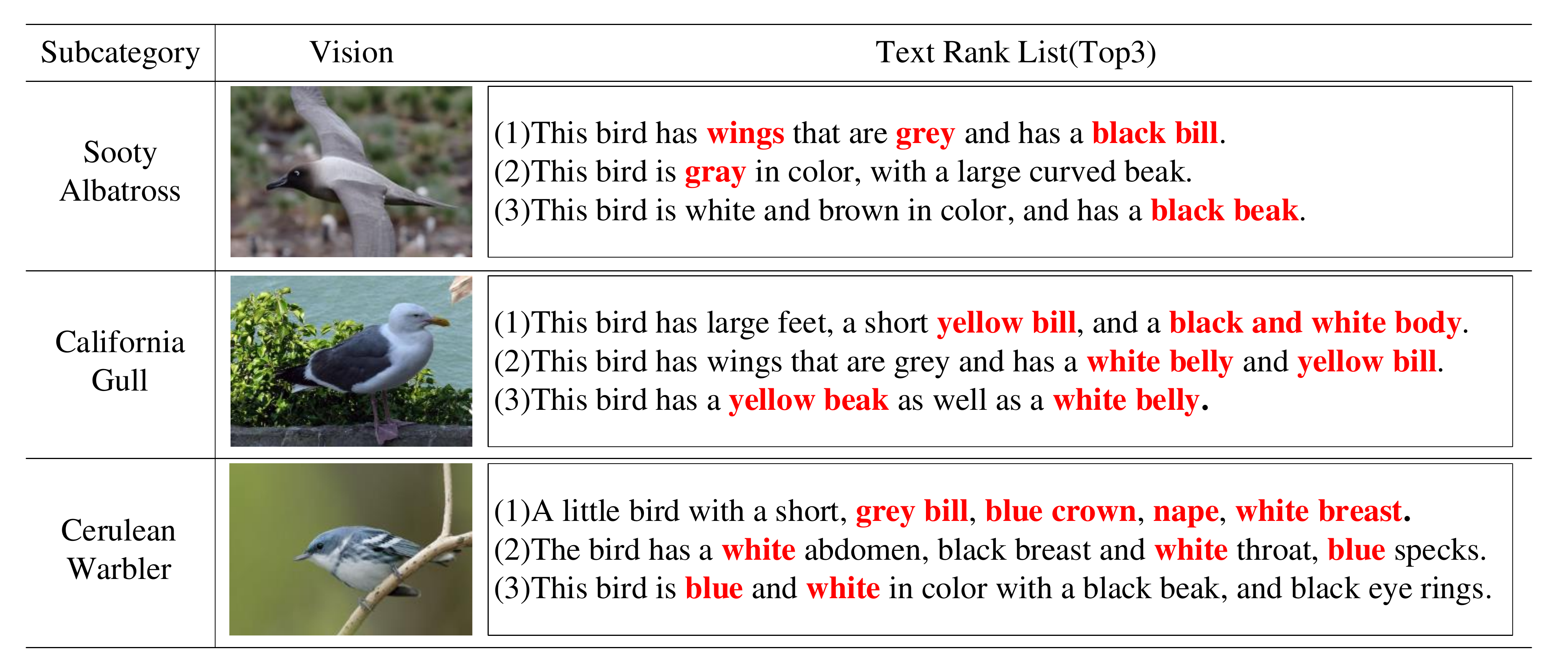}
\end{center}
   \caption{Some results of the textual stream. }
\label{visonandlanguage}
\end{figure}

\subsubsection{Effectiveness of Visual-textual Representation Learning}
\begin{figure}[!t]
\begin{center}
\includegraphics[width=1\linewidth]{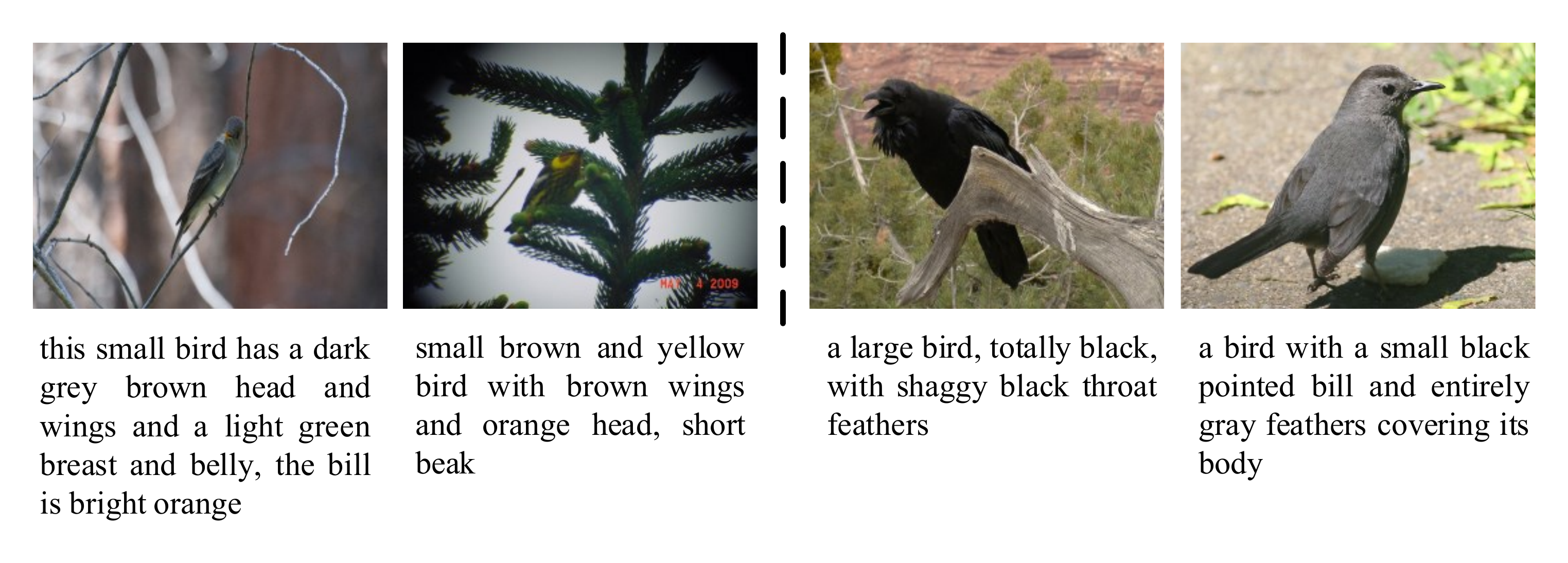}
\end{center}
   \caption{Some example results where the textual and visual streams are complementary. The left two images are rightly categorized by textual stream, but wrongly categorized by visual stream. The right two images are just the opposite.}
\label{difference}
\end{figure}

We also present the baseline experiment to verify the effectiveness of visual-textual representation learning. The results are shown in Table \ref{cvl}, where ``VTRL-textual'' means textual stream, ``VTRL-visual'' means visual stream and ``VTRL(only original image'' means only a fine-tuned CNN model is used. We can observe that categorization result of textual stream is promising. From the first line of each row in Fig. \ref{visonandlanguage}, we can find that textual description with the highest score always points out the discriminative characteristics of the object, as the red words shows. 
Combining visual and textual information can further achieve more accurate categorization result, which demonstrates that the two types of information are complementary: visual information focuses on the global and local features, and textual information further points the importance of these features.
Fig. \ref{difference} shows some example results where the textual and visual streams are complementary.	Visual stream is effective for dealing with those images, which have few discriminative characteristics. Humans can only describe them in a rough way, but cannot describe them in detail, which leads that the textual information carries less useful information to distinguish it from other subcategories. Examples are shown as the right two images.
Textual stream is effective for dealing with those images, whose foreground and background are hard to be distinguished by visual stream. But they can be described in details by text, which carries the information of the discriminative characteristics and be helpful for the categorization. Examples are shown as the left two images.

Besides, we also compare our VTRL approach with method based on both textual and visual attention, such as Co-attention \cite{lu2016hierarchical}. It only achieves the accuracy of 73.90\%, which is lower than our VTRL approach. It is mainly because that our VTRL approach discovers the fine-grained visual-textual patterns, which are key hints to the fine-grained categorization.

From the above baseline results, the separate contribution of each component in our proposed VTRL approach can be verified. First, object localization and fine-grained pattern mining discover the discriminative and representative information of image via visual-textual attention.
Second, the complementarity between visual and textual information is fully captured by visual-textual representation learning.

\section{Conclusions}
\label{conclusion}
In this paper, the fine-grained visual-textual representation learning approach has been proposed.  
Based on textual attention, we employ fine-grained visual-textual pattern mining to discover discriminative information for categorization through jointly modeling vision and text with GANs.  
Then, visual-textual representation learning jointly considers visual and textual information, which preserves the intra-modality and inter-modality information to generate complementary fine-grained representation, and further improve categorization performance. Experimental results on two widely-used fine-grained visual categorization datasets demonstrate the superiority of our approach compared with state-of-the-art methods. 

As for the future work, we will focus on the following two aspects: First, we will attempt to extend the current two-stream framework into an end-to-end framework for simplifying the process. Second, we will exploit exact and effective methods on relating textual attention and visual attention for more accurate discriminative parts localization as well as better categorization performance.




\ifCLASSOPTIONcaptionsoff
  \newpage
\fi



%
\balance{
\bibliographystyle{plain}
\bibliography{reference}
}

%








\end{document}